\documentclass[lettersize,journal]{IEEEtran}
\usepackage{amsmath,amsfonts}
\usepackage{algorithmic}
\usepackage{algorithm}
\usepackage{array}
\usepackage[caption=false,font=normalsize,labelfont=sf,textfont=sf]{subfig}
\usepackage{textcomp}
\usepackage{stfloats}
\usepackage{url}
\usepackage{verbatim}
\usepackage{graphicx}
\usepackage{cite}
\usepackage{siunitx}
\usepackage{multirow}
\usepackage{booktabs}
\usepackage{bm}

\begin{document}

\title{From Instance Segmentation to 3D Growth Trajectory Reconstruction in Planktonic Foraminifera}
\author{Huahua Lin, Xiaohao Cai, Mark Nixon, James M. Mulqueeney, Thomas H. G. Ezard}



\maketitle

\begin{abstract}
Planktonic foraminifera, marine protists characterized by their intricate chambered shells, serve as valuable indicators of past and present environmental conditions. Understanding their chamber growth trajectory provides crucial insights into organismal development and ecological adaptation under changing environments. However, automated tracing of chamber growth from imaging data remains largely unexplored, with existing approaches relying heavily on manual segmentation of each chamber, which is time-consuming and subjective. In this study, we propose an end-to-end pipeline that integrates instance segmentation, a computer vision technique not extensively explored in foraminifera, with a dedicated chamber ordering algorithm to automatically reconstruct three-dimensional growth trajectories from high-resolution computed tomography scans. We quantitatively and qualitatively evaluate multiple instance segmentation methods, each optimized for distinct spatial features of the chambers, and examine their downstream influence on growth-order reconstruction accuracy. Experimental results on expert-annotated datasets demonstrate that the proposed pipeline substantially reduces manual effort while maintaining biologically meaningful accuracy. Although segmentation models exhibit under-segmentation in smaller chambers due to reduced voxel fidelity and subtle inter-chamber connectivity, the chamber-ordering algorithm remains robust, achieving consistent reconstruction of developmental trajectories even under partial segmentation. This work provides the first fully automated and reproducible pipeline for digital foraminiferal growth analysis, establishing a foundation for large-scale, data-driven ecological studies.
\end{abstract}


\section{Introduction}
\IEEEPARstart{P}{lanktonic} foraminifera are unicellular marine organisms that construct calcareous shells, or tests, composed of multiple chambers added in a sequential pattern as the organism grows. These chambers follow species-specific arrangements that serve as critical indicators for taxonomic classification and evolutionary adaptation \cite{brummer2022taxonomic}. The spatial and temporal sequence in which chambers are added—referred to as chamber ordering—is a fundamental morphological characteristic that provides insight into developmental processes and life history strategies \cite{kucera2007chapter} as well as the assignment of individuals to species \cite{aze2011phylogeny} and thus questions about the processes by which biodiversity generates, proliferates and eradicates.

With the advent of high-resolution computed tomography (CT), detailed cross-sectional imaging of foraminiferal chamber structures has become possible, enabling the reconstruction of their three-dimensional (3D) growth trajectories \cite{speijer2008quantifying}. Despite the increasing availability of such volumetric data, there remains a lack of automated tools and annotated datasets specifically designed for the morphological analysis of planktonic foraminifera \cite{he2024opportunities}. As a result, the study of chamber ordering continues to rely heavily on manual workflows. Existing workflows require researchers to manually identify and label individual chambers either across 2D slices \cite{schmidt2013linking, caromel2016morphological} or in full 3D reconstructions \cite{burke2020three}. Once the chambers are labeled, geometric features such as volume and spatial centroid can be quantified to reconstruct the growth trajectory \cite{brombacher2023developmental}. However, this manual process is time-consuming and subjective, particularly when dealing with large datasets \cite{joskowicz2019inter}.

While recent efforts have advanced automated classification of planktonic foraminifera using deep learning \cite{hsiang2019endless, mitra2019automated}, automated chamber segmentation and ordering remain significantly underexplored. Previous studies by Ge et al. \cite{ge2017coarse, ge2021enhancing} have investigated chamber segmentation using edge-based methods; however, these approaches are based on 2D microscope images captured under varying lighting orientations, and are not directly applicable to volumetric CT data. Brombacher et al. \cite{brombacher2022analysing} proposed an ordering approach based on geometric priors such as centroid distances and angles, but their approach still requires manual chamber delineation as a prerequisite. Furthermore, the compact arrangement and partial connectivity of chambers often obscure clear separations in CT data, making both manual and automated segmentation inherently challenging.

To address these gaps, we build upon the recently released dataset \cite{brombacher2025detecting}, which comprises 50 high-resolution CT-scanned specimens of \textit{Menardella}, with all chambers annotated at the instance level. We propose the first end-to-end pipeline for automated chamber segmentation and growth-order reconstruction in planktonic foraminifera. Our method eliminates manual intervention and allows scalable and objective morphological analysis. The proposed pipeline begins with an instance segmentation stage that combines a deep learning-based semantic segmentation module with hand-crafted post-processing to produce instance-level chamber labels. The semantic segmentation step identifies Regions of Interest (ROIs) such as chamber interiors, chamber boundaries, and background areas. From these segmented regions, geometric properties including spatial locations and volumes of individual chambers are extracted. These features are then used to infer the sequential growth order through a nearest-neighbor-based ordering algorithm \cite{rosenkrantz1977analysis}. We benchmark multiple mainstream segmentation methods, including both 2D and 3D configurations with varying ROI definitions, and assess their effectiveness in supporting downstream growth trajectory reconstruction.

In summary, our main contributions include:
\begin{itemize}
    \item We propose an end-to-end pipeline for automated chamber ordering that integrates instance segmentation with a dedicated ordering algorithm. The proposed pipeline enables large-scale, objective morphological analysis by eliminating manual annotation in the ordering process.
    \item We present the first systematic adaptation and benchmarking of mainstream instance segmentation methods in biological image tasks for CT scans of the \textit{Menardella} specimens. The evaluated methods demonstrate strong generalizability and hold potential for analyzing other chamber-secreting organisms.
    \item We incorporate a nearest-neighbor-based ordering algorithm to infer 3D chamber growth trajectory from segmented volumes. The resulting orderings exhibit a high Spearman correlation with the ground-truth sequences, validating the accuracy and robustness of the proposed pipeline.
\end{itemize}

The remainder of the paper is organized as follows. Section~\ref{sec:related_works} reviews common instance segmentation approaches, including hand-crafted, deep learning, and hybrid methods. Section~\ref{sec:method} details the proposed pipeline developed for the final objective of chamber ordering. Section~\ref{sec:data} describes the dataset of the \textit{Menardella} genus. Sections~\ref{sec:exp} and~\ref{sec:discussion} present the experimental results and provide a comparative analysis of the different segmentation methods. Finally, Section~\ref{sec:conclusion} concludes the paper.

\section{Related Works}
\label{sec:related_works}
Accurate identification of individual chambers is fundamental to the analysis of growth patterns in planktonic foraminifera. Automating this task requires instance segmentation, a computer vision technique that delineates each instance within a 2D image or a 3D volume. Over time, methodological approaches to instance segmentation have evolved from traditional mathematical algorithms to modern deep learning-based frameworks, each offering distinct strengths and limitations for chamber segmentation.

\subsection{Hand-Crafted Methods}
Traditional region-based segmentation techniques operate by grouping pixels into coherent regions based on homogeneity criteria such as intensity, texture, or other local features. The watershed transform \cite{vincent1991watersheds} stands as one of the most well-known examples in this category, treating image intensity as a topographic surface and flooding basins from seed regions such as intensity minima or user-provided markers. However, the watershed approach suffers from over-segmentation due to sensitivity to noise and local minima, prompting the development of marker-controlled variants \cite{meyer1990morphological, volkmann2002novel, straehle2012seeded}. Beyond watershed, region growing methods provide an alternative strategy by iteratively expanding seed points to include neighboring pixels that satisfy predefined similarity criteria \cite{adams1994seeded}. For automated applications, several extensions combine automatic seed selection with adaptive similarity measures, which have proven effective in complex imaging contexts such as color image segmentation \cite{shih2005automatic}.

Clustering-based methods offer a complementary perspective by treating segmentation as a feature space grouping problem. Graph-based clustering techniques partition images by treating pixels as nodes and defining edge weights based on feature similarity, with some variants employing signed graphs to encode both attraction and repulsion relationships between regions \cite{bailoni2022gasp, chiang2012scalable, kunegis2010spectral}. However, these methods typically require supernode pre-processing to remain computationally tractable. Mean-shift clustering provides another option, operating in feature space to iteratively shift candidate regions toward local density maxima without requiring preset cluster counts \cite{comaniciu2002mean}.

\subsection{Deep Learning Methods}
A comprehensive survey of deep learning-based instance segmentation methods is presented in \cite{hafiz2020survey}, highlighting the evolution of approaches in this field. The dominant paradigm involves a two-stage process combining object detection with mask prediction, where bounding box proposals are first generated and subsequently segmented \cite{he2017mask, cai2019cascade, liu2018path, chen2019tensormask}. Mask R-CNN (Region-based Convolutional Neural Network) \cite{he2017mask} exemplifies this approach, employing a region proposal network (RPN) to identify candidate ROIs before performing classification and mask prediction in the second stage. However, the segmentation accuracy depends on precise feature localization, and the sequential processing pipeline significantly increases inference time. 

In contrast, single-stage approaches like YOLACT \cite{bolya2019yolact} offer a more efficient alternative by decoupling the segmentation process. This architecture simultaneously generates a set of prototype masks and instance-specific coefficients, with final instance masks produced through a linear combination of these components. This design achieves competitive accuracy with substantially higher throughput, mitigating the computational bottlenecks inherent in two-stage frameworks.

More recently, the segment anything model (SAM) \cite{kirillov2023segment} introduces a new direction by framing segmentation as a promptable vision task, enabling zero-shot generalization to unseen objects and domains. Instead of relying on bounding box proposals, SAM uses prompts such as points or boxes to generate segmentation masks directly. While SAM is a breakthrough in zero-shot segmentation, its performance on domain-specific data remains limited without additional fine-tuning or domain adaptation.

\subsection{Hybrid Methods}
The aforementioned deep learning methods are primarily designed for instance segmentation on 2D images. While there are advanced instance segmentation approaches tailored for 3D data \cite{schult2023mask3d, chen2021hierarchical, hou20193d, wang2018sgpn}, most are developed for point cloud representations commonly used in scene understanding tasks such as autonomous driving or robotics. In contrast, volumetric biological data—such as that obtained from CT or MRI scans—presents a different set of challenges, including dense, homogeneous textures, anisotropic resolution, and the need for high precision in segmenting tightly packed structures.

In the domain of 3D biological imaging, the prevailing workflow still relies heavily on 3D deep learning models for semantic segmentation, followed by traditional post-processing techniques to achieve instance-level delineation. Several hybrid frameworks for instance segmentation have been actively explored \cite{kar2022benchmarking, wang2022novel, zhao2021faster, wolny2020accurate}, leveraging the representational power of deep learning and the interpretability of traditional methods. For example, deep watershed transform \cite{bai2017deep} trains a CNN to predict energy maps that guide a watershed-based instance segmentation, while InstanceCut \cite{kirillov2017instancecut} employs a CNN to perform semantic segmentation and instance-aware edge detection, which are then combined through a MultiCut framework to resolve instance groupings. These hybrid strategies demonstrate how coupling deep feature representations with topological or graph-based post-processing can enhance segmentation precision.

\section{Method}
\label{sec:method}
\subsection{Pipeline Overview}
\begin{figure*}[!htbp]
    \centering
        \includegraphics[width=\linewidth]{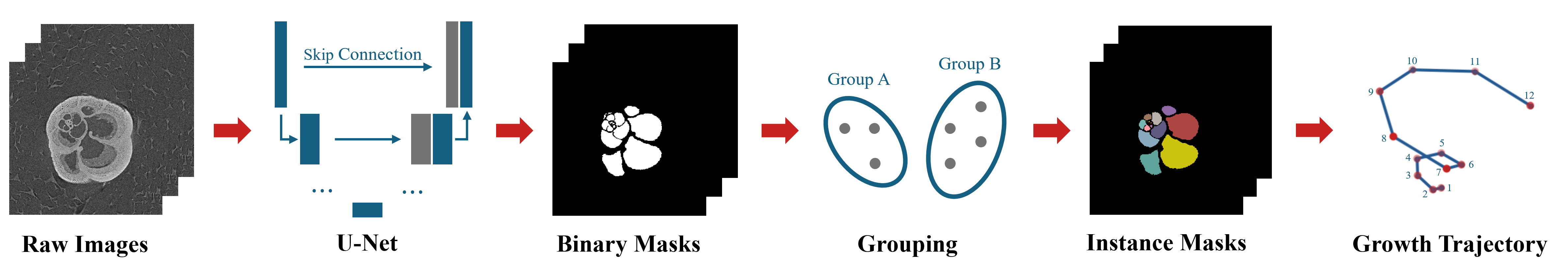}
    \caption{An overview of the developed pipeline for determining chamber growth trajectories in planktonic foraminifera. The process starts with raw CT scan images, which pass through a deep neural architecture like U-Net to produce binary masks that identify chamber regions. A grouping step using hand-crafted methods follows to separate individual chambers, resulting in instance masks. Spatial features extracted from these masks support the inference of the 3D chamber growth trajectory.}
    \label{fig:pipeline}
\end{figure*}

As depicted in Fig. \ref{fig:pipeline}, the developed pipeline begins with raw CT scans that are first downsampled for computational efficiency and then fed into a deep neural architecture for semantic segmentation of the ROIs, from which binary masks of chambers are derived. We adopt the standard U-Net model as our backbone, as alternative architectures such as UNet++ \cite{zhou2019unet++} and SwinUNETR \cite{hatamizadeh2021swin} yielded similar performance on our dataset. U-Net thus provides a lightweight yet sufficiently expressive model for segmentation.

Subsequently, these binary masks are refined using traditional grouping techniques to achieve precise instance masks of individual chambers. We evaluate multiple segmentation variants (see Section~\ref{subsec:seg-method}), each targeting distinct ROIs and employing tailored grouping procedures. Once individual chambers are identified, their spatial information is extracted and used to infer 3D growth trajectories for each specimen (see Section \ref{subsec:ordering-method}), revealing the sequence of chamber addition during ontogeny.

\subsection{Segmentation Methods for 3D Chamber Delineation}
\label{subsec:seg-method}
\subsubsection{U-Net + SW}
The first segmentation method is based on the standard 2D U-Net \cite{ronneberger2015u} or its 3D extension \cite{cciccek20163d}, comprising an encoder–decoder structure designed for binary segmentation to delineate the chamber interiors from the background. Each level of the encoder consists of two convolutional layers with a kernel size of 3, each followed by a rectified linear unit (ReLU) activation and batch normalization, and concludes with a max pooling operation with a stride of 2 for downsampling. The number of feature channels doubles after each downsampling step. The decoder mirrors this structure, with each level beginning with an upsampling operation via transposed convolution, followed by two convolutional layers with ReLU activation and batch normalization. Skip connections concatenate the corresponding encoder and decoder features at equivalent hierarchical levels. 

We train U-Nets by using the Binary Cross-Entropy (BCE) loss. For an input 2D image $\mathbf{I}\in\mathbb{R}^{h\times w}$ or 3D volume $\mathbf{I}\in\mathbb{R}^{h\times w \times d}$, the dimensions $h$, $w$, and $d$ denote height, width, and depth (number of slices), respectively. Let $\mathbf{G}\in\{0,1\}^N$ and $\mathbf{P}\in[0,1]^N$ denote the ground truth labels and predicted probabilities of an input, respectively, where $N$ is the total number of pixels or voxels. Here, the $i$-th element of $\mathbf{G}$ say $g_i$ with $g_i=1$ represents the chamber interiors, and $g_i=0$ denotes the background. With $p_i\in[0,1]$ being the network's predicted probability at location $i$, the BCE loss averaged over all spatial locations is then defined as
\begin{equation}
    \mathcal{L}_{\text{BCE}} = -\frac{1}{N}\sum_{i=1}^N\left(g_i\log(p_i) + (1-g_i)\log(1-p_i)\right).
\end{equation}

A threshold $\tau=0.5$ is applied to obtain binary semantic masks, which are subsequently refined using the Seeded Watershed (SW) algorithm to segment individual chambers. The SW method is applied to the 3D volume, and in the 2D case, predictions are stacked to form this volume. Direct watershed on the distance transform (i.e., each pixel in the foreground is assigned a value equal to its distance to the nearest background pixel) often causes over-segmentation due to the close proximity between chambers. To alleviate this, morphological erosion \cite{cuisenaire1999fast} is used to isolate confident chamber interiors, effectively separating adjacent chambers while preserving the maximum number of small chambers. Connected Component Labeling (CCL) \cite{6291402} with a 6-connected neighborhood is then applied to identify distinct regions, with small noisy components removed. The resulting labels serve as watershed markers, enabling accurate chamber separation while mitigating over-segmentation.

\subsubsection{3D PlantSeg}
The second segmentation method adapts the PlantSeg framework \cite{wolny2020accurate}, originally designed for 3D plant cell instance segmentation, leveraging its effectiveness for the analogous task of segmenting foraminifera chambers. Like plant cells, foraminiferal chambers occupy discrete spatial domains with clear boundary contrasts in micro-CT data, making boundary-aware graph partitioning suitable for this application.

Specifically, a modified 3D U-Net (see \cite{wolny2020accurate} for details) is trained to predict boundary probability maps using a loss function combining BCE and Dice loss, i.e.,
\begin{equation}
    \mathcal{L}_{\text{PlantSeg}} = \mathcal{L}_{\text{BCE}} + \mathcal{L}_\text{{Dice}}, 
\end{equation}
with $\mathcal{L}_{\text{Dice}}$ given by
\begin{equation}
    \mathcal{L}_{\text{Dice}} = 1 - \frac{2\sum_{i=1}^Np_ig_i}{\sum_{i=1}^Np_i^2 + \sum_{i=1}^Ng_i^2}.
\end{equation}

After predicting chamber boundaries, watershed clustering is applied to the distance transform of the thresholded boundary map ($\tau=0.3$) to generate supervoxels. These supervoxels form nodes of a Region Adjacency Graph (RAG), where edges connect spatially adjacent regions and edge weights correspond to the mean boundary probabilities along shared interfaces. The resulting signed graph is partitioned using the Generalized Algorithm for Signed Graph Partitioning (GASP) \cite{bailoni2022gasp} with a merge affinity of 0.8. This optimization jointly enforces spatial continuity and boundary consistency, merging over-segmented regions while preserving high-confidence boundaries.

\subsubsection{3D MTL + SW}
While the previous segmentation methods focus on either chamber interiors or boundaries, these tasks are inherently complementary: interior segmentation captures volumetric occupancy, whereas boundary detection enhances localization precision. To leverage both types of information, we adopt a Multi-Task Learning (MTL) framework that jointly predicts 3D chamber interiors, chamber boundaries, and background within a unified model.

The framework processes volumetric images $\mathbf{I}\in\mathbb{R}^{h\times w \times d}$ using a shared 3D U-Net encoder $E(\cdot)$ and three decoder branches $\{D_c(\cdot)\}_{c=1}^3$, each producing voxel-wise probability maps $\mathbf{P}_c=D_c(E(\mathbf{I}))\in\left[0,1\right]^{N}$ for class $c$ (interior, boundary, or background). The overall loss sums the weighted focal losses across all categories along with a consistency loss defined as
\begin{equation}
    \mathcal{L}_{\text{MLT}} = \sum_{c=1}^3\lambda_c\mathcal{L}_{\text{Focal}}^{(c)} + \mathcal{L}_{\text{Consistency}},
\end{equation}
where $\lambda_c$ is the balancing coefficients, each category-specific $\mathcal{L}_{\text{Focal}}^{(c)}$ \cite{lin2017focal} is expressed by
\begin{equation}
    \mathcal{L}_{\text{Focal}}^{(c)} = -\frac{1}{N}\sum_{i=1}^{N}\alpha_c(1-p_{c,i})^{\gamma_c}\log(p_{c,i}),
\end{equation}
with $\alpha_c$ and $\gamma_c$ being the class weights and focusing parameter, respectively,
and the consistency regularization term $\mathcal{L}_{\text{Consistency}}$ is formulated as
\begin{equation}
    \mathcal{L}_{\text{Consistency}} = -\frac{1}{N}\sum_{i=1}^N\Big(\log\big(\sum_{c=1}^3p_{c,i}\big)\Big).
\end{equation}
The consistency term enforces mutual exclusivity among categories, encouraging the three probability maps to sum to 1 at each voxel and thus reducing label ambiguity.

The weighting parameters $\alpha_c$, $\gamma_c$, and $\lambda_c$ were empirically tuned to emphasize boundary voxels while preventing background dominance, following standard focal loss practices \cite{lin2017focal}. The final chamber mask is constructed by thresholding the three probability maps ($\tau_1=0.5$, $\tau_2=0.4$, $\tau_3=0.9$) and retaining interior voxels not overlapping with boundary or background regions. CCL with a 6-connected neighborhood is then applied to the smoothed mask to identify individual chamber seeds, excluding small and noisy components. Finally, the SW algorithm operates on the non-background mask guided by these markers, producing instance-level segmentation that preserves structural coherence and resolves weakly connected chambers.

\subsection{Ordering of 3D Chambers}
\label{subsec:ordering-method}
The labels generated by the segmentation methods do not inherently encode any ordering information among the chambers. To recover the growth trajectory, we further apply a nearest-neighbor algorithm, i.e., summarized in Algorithm \ref{alg:nn} (with the notation used given below), to infer the sequential arrangement from the segmented chambers. This approach is motivated by the biological growth pattern of \textit{Menardella}, where chambers are added sequentially in a spatially adjacent manner. The chamber with the smallest volume is assumed to be the initial one, as it corresponds to the earliest ontogenetic stage.

\textit{Notation.}  Let $\mathcal{S}=\{S_1, S_2, \dots, S_M\}$ be a set of labeled chambers, where $M$ is the total number of chambers and each $S_j$ represents an individual chamber indexed by $j\in\{1,2,\dots,M\}$. The chambers ordering algorithm in Algorithm~\ref{alg:nn}  first computes the volume $V(S_j)$ and centroid $C(S_j)$ of each chamber $S_j\in \mathcal{S}$, and then initializes the ordered sequence $\mathcal{T}$ with the chamber of minimum volume $S_{\rm start}$. At each step, the algorithm adds to $\mathcal{T}$ the remaining unvisited chamber stored in $\mathcal{U}$ whose centroid is closest to the current one $c_{\rm current}$ in terms of Euclidean distance. This iterative process continues until all chambers are included in $\mathcal{T}$, resulting in a complete, biologically plausible growth path.

\begin{algorithm}
\caption{Nearest-Neighbor-based Chamber Growth Path}
\label{alg:nn}
\begin{algorithmic}[1]
\REQUIRE Labeled chambers $\mathcal{S} = \{S_1,S_2, \dots, S_M\}$
\ENSURE Ordered list $\mathcal{T}$ representing the growth path

\STATE Compute volumes $V(S_j)$ and centroids $C(S_j)$ for all chambers $S_j \in \mathcal{S}$
\STATE $S_{\text{start}} \gets \arg\min_{S_j \in \mathcal{S}} \{V(S_j)\}$
\STATE $\mathcal{T} \gets [S_{\text{start}}]$
\STATE $\mathcal{U} \gets \mathcal{S} \setminus \{S_{\text{start}}\}$
\STATE $c_{\text{current}} \gets C(S_{\text{start}})$
\WHILE{$\mathcal{U} \neq \emptyset$}
    \STATE $S_{\text{next}} \gets \arg\min_{S_j \in \mathcal{U}} \{
    \left\lVert C(S_j) - c_{\text{current}} \right\rVert_2 \}$
    \STATE Append $S_{\text{next}}$ to $\mathcal{T}$
    \STATE $c_{\text{current}} \gets C(S_{\text{next}})$
    \STATE $\mathcal{U} \gets \mathcal{U} \setminus \{S_{\text{next}}\}$
\ENDWHILE
\RETURN $\mathcal{T}$

\end{algorithmic}
\end{algorithm}

\section{Data} \label{sec:data}
\begin{figure*}[!htbp]
    \centering
        \includegraphics[width=0.9\linewidth]{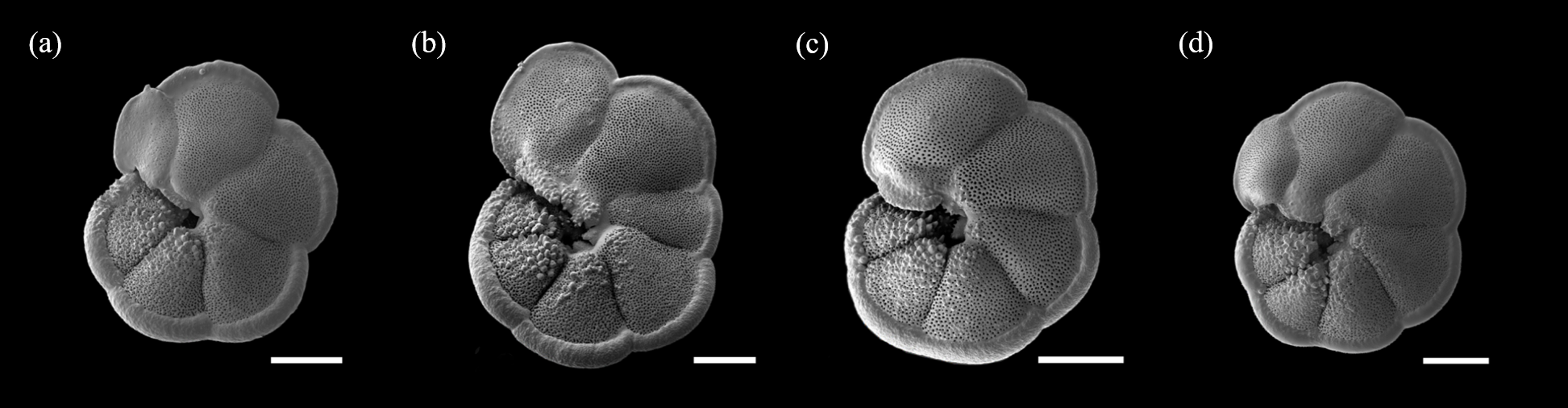}
    \caption{Example images of the four different species of planktonic foraminifera \cite{woodhouse2023paleoecology} used in our dataset. (a) \textit{Menardella exilis}, (b) \textit{Menardella limbata}, (c) \textit{Menardella menardii} and (d) \textit{Menardella pertenuis}. The scalar bar for images is 200 microns.}
    \label{fig:species}
\end{figure*}

Advancements in high-resolution X-ray CT now allow for detailed observation and analysis of both internal and external structures of planktonic foraminifera, which are less than 1mm in diameter. In contrast to 2D optical microscopy datasets such as Endless Forams \cite{hsiang2019endless}, which capture only surface morphology, X-ray CT provides volumetric data that reveal internal chamber organization and spatial relationships essential for developmental analysis. In this study, we utilize an existing dataset of 50 specimens of the genus \textit{Menardella} \cite{brombacher2025detecting}, covering 4 different species: \textit{Menardella exilis}, \textit{Menardella limbata}, \textit{Menardella menardii} and \textit{Menardella pertenuis}. Example images are shown in Fig.~\ref{fig:species}.

\begin{figure*}[!htbp]
\centering
\subfloat[]{\includegraphics[width=0.45\linewidth]{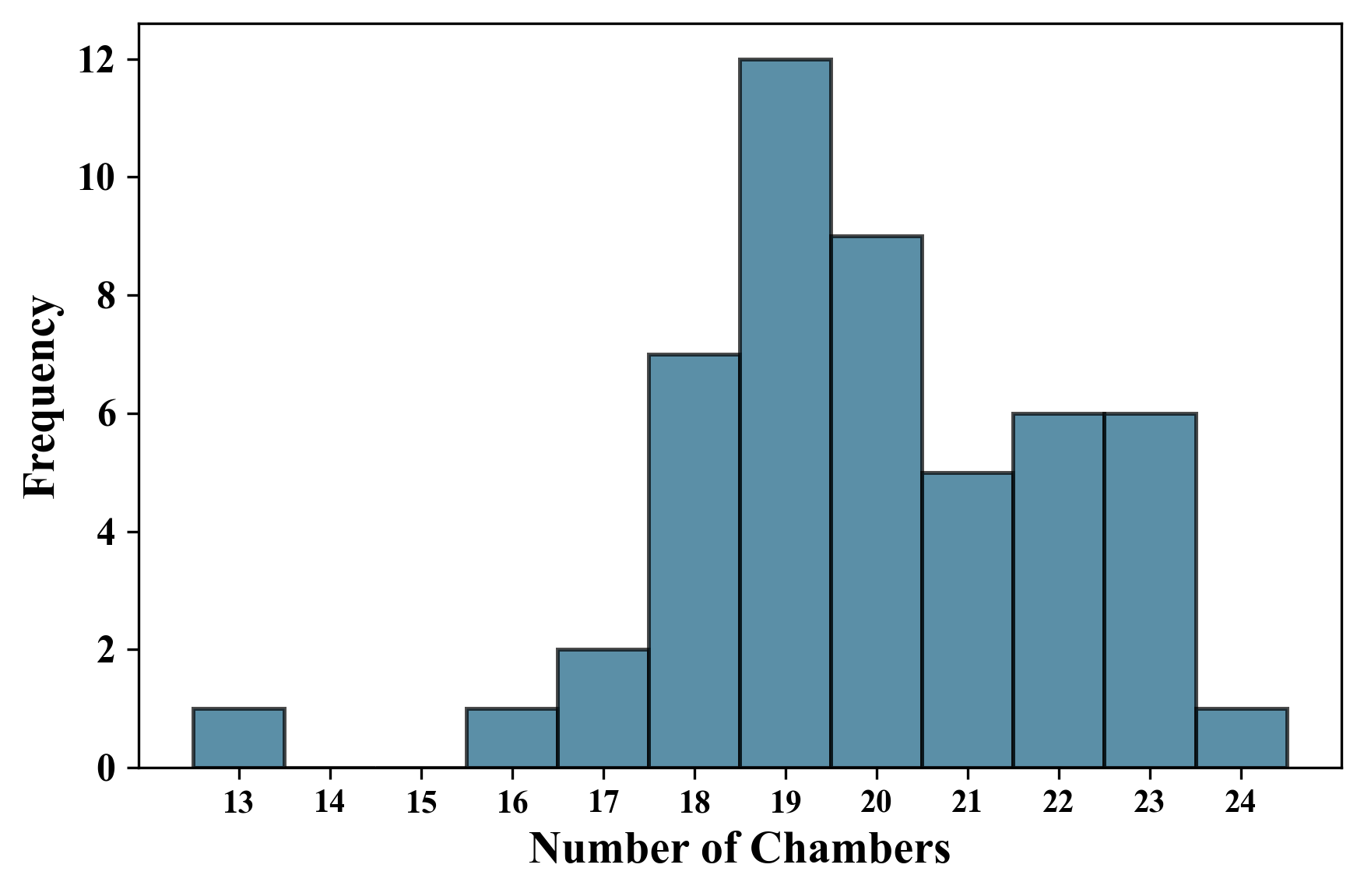}%
\label{distribution_num_chambers}}
\hfil
\subfloat[]{\includegraphics[width=0.45\linewidth]{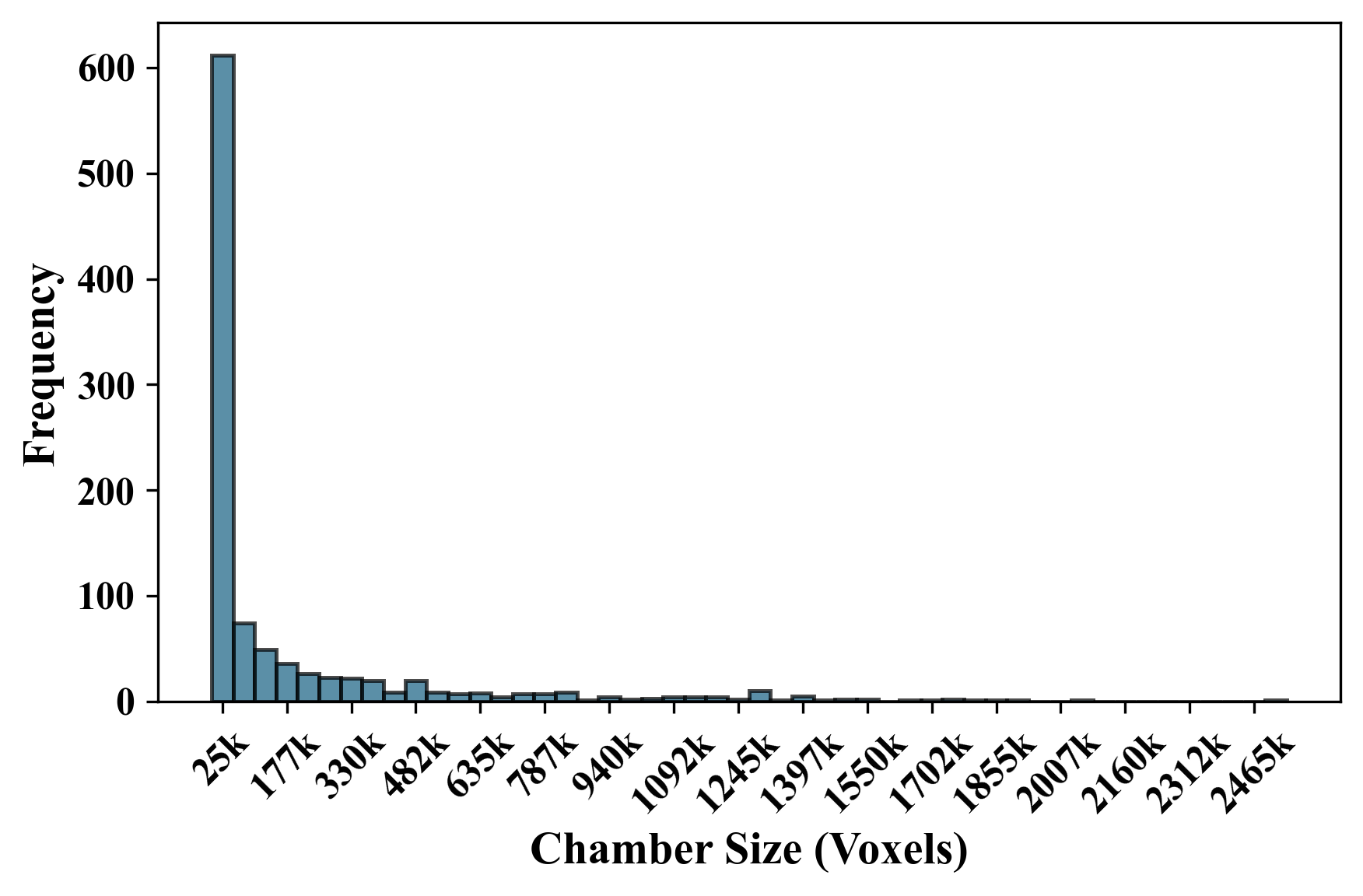}%
\label{distribution_chamber_size}}
\caption{Statistical distributions of chamber characteristics in our dataset. (a) Histogram of chamber counts per specimen (total $50$ specimens) showing a unimodal distribution (range: 13--24 chambers). (b) Logarithmic distribution of individual chamber sizes (total $1024$ measurements) revealing heavy-tailed morphology spanning $10^2$--$10^6$ voxels, indicating right-skewed size variation.}
\label{fig:datasets_desc}
\end{figure*}

Based on key features such as chamber size and apertures position, each sample is labeled with a sequence of numbers starting from 1, denoting the chronological growth order of chambers, while label 0 represents the background. The dataset exhibits natural variability in chamber counts, ranging from 13 to 24 chambers (see the distribution in Fig.~\ref{fig:datasets_desc}a), which reflects differences in ontogenetic growth stages and environmental conditions. Each sample comprises 120 to 600 2D slices, each with a resolution of 992 $\times$ 1015 pixels, capturing chamber volumes spanning four orders of magnitude ($10^2$--$10^6$ voxels; Fig.~\ref{fig:datasets_desc}b). This heterogeneity is driven by exponential scaling in later growth stages. Further details on the dataset can be found in \cite{brombacher2025detecting, mulqueeney2024many}. For machine learning applications, the data were partitioned into 70\% training and 30\% test sets.

\section{Experiments and Results}
\label{sec:exp}
This section presents the experimental evaluation of the proposed pipeline for chamber ordering in the genus \textit{Menardella} specimens. All segmentation methods are assessed using a set of quantitative metrics to evaluate both segmentation quality and the accuracy of the reconstructed chamber sequences, along with qualitative visualization for interpretability. Comparative results across all segmentation methods are reported, with further analysis provided in Section \ref{sec:discussion}.

\subsection{Implementation Details}
\subsubsection{U-Net + SW}
All input images were cropped and resized to a resolution of 256$\times$256 ($h\times w$) for 2D and 128$\times$128$\times$64 ($h\times w\times d$) for 3D before being fed into their respective models. Data augmentation for both 2D and 3D cases includes random flips and 90-degree random rotations along each axis. The ground truth of chamber interiors was created by converting the annotations into a binary format. Specifically, all pixels with a label greater than 0, indicating the presence of a chamber, are assigned a value of 1, while pixels labeled as 0, representing the background, are assigned a value of 0.

Model optimization was performed using the Adam optimizer with a learning rate of 1E$-3$ for the 2D U-Net, while the 3D U-Net was trained using Stochastic Gradient Descent (SGD) with a learning rate of 1E$-2$, momentum of 0.9, and a weight decay of 5E$-4$. The mini-batch sizes were set to 32 for 2D and 4 for 3D. Training until convergence required a minimum of 2 hours for the 2D U-Net on a single NVIDIA A100 80GB GPU over 50 epochs, and approximately 1.5 hours for the 3D U-Net over 100 epochs using 4 GPUs.

Post-processing steps for instance segmentation—including erosion, CCL, and SW—were directly applied to the 3D volumetric data, which could be obtained either from the output of a 3D U-Net or by stacking the predicted slices generated by a 2D U-Net. When using the stacked 2D predictions, the pipeline required approximately 2.34 seconds per sample, compared to 0.12 seconds for the 3D U-Net output, due to the higher spatial resolution of the 2D predictions.

\subsubsection{3D PlantSeg}
All volumetric input images were preprocessed to a fixed resolution of 128$\times$128$\times$64 voxels ($h\times w\times d$) through cropping and resizing before network input. Intensity normalization and random flips along each axis were used for data augmentation. Since ground truth annotations for chamber boundaries (walls) are not explicitly provided, we derived them by computing the distance transform on each ground truth binary mask. Given that the annotated chambers are either well separated or minimally touching, the distance transform effectively highlighted inter-chamber boundaries. Voxels with distance values less than or equal to 1 were designated as ground truth boundaries for each sample.

For optimization, we used the Adam optimizer with a learning rate of 1E$-3$ and a mini-batch size of 2. Training was conducted on 4 NVIDIA A100 GPUs, requiring approximately 3 hours to converge over 200 epochs. The graph partitioning process, including the watershed algorithm and GASP, required approximately 0.52 seconds per sample.

\subsubsection{3D MTL + SW}
Before model training, all input images were cropped and resampled to a resolution of 128$\times$128$\times$64 voxels ($h\times w\times d$). Random flips were performed along each spatial axis to improve model generalization. The boundary labels matched those used in PlantSeg's training process, while the interior chamber labels were refined by subtracting boundary voxels from the ground truth labels used in U-Net + SW, and the background labels were computed as the logical negation of the chamber interior and boundary masks.

The optimization settings followed those used in PlantSeg. Although the model simultaneously predicted three output categories, the overall training time was comparable to the other 3D networks, indicating efficient multi-task convergence. The post-processing steps, including CCL and SW, took around 0.01 seconds for each sample, demonstrating minimal runtime overhead.

\subsection{Evaluation Metrics}
\subsubsection{Segmentation Accuracy}
We evaluated segmentation quality using the Intersection over Union (IoU) \cite{everingham2010pascal} and the Adjusted Rand Index (ARI) \cite{hubert1985comparing}. IoU measures the spatial overlap between predicted and ground truth masks, focusing on the accuracy of shape alignment. In contrast, the ARI quantifies the similarity between predicted and ground truth segmentations by evaluating all pairs of voxels, measuring how consistently voxel pairs are assigned to the same or different instances. It emphasizes the correctness of instance grouping regardless of label permutations, and penalizes over- and under-segmentation.

For a predicted segmentation $\mathbf{P}$ and ground truth $\mathbf{G}$, where both $\mathbf{P}$ and $\mathbf{G}$ represent 3D volumes formed either by stacking 2D slices or directly as 3D data, the IoU is defined as
\begin{equation}
    \text{IoU}(\mathbf{P},\mathbf{G}) = \frac{|\mathbf{P}\cap \mathbf{G}|}{|\mathbf{P}\cup \mathbf{G}|},
\end{equation}
with values ranging from 0 (no overlap) to 1 (perfect overlap).
Whilst the ARI is derived from the Rand Index (RI) \cite{rand1971objective} by accounting for chance agreement, i.e.,
\begin{equation}
\text{ARI}(\mathbf{P},\mathbf{G}) = \frac{\text{RI}(\mathbf{P},\mathbf{G}) - E[\text{RI}]}{1 - E[\text{RI}]},
\end{equation}
where $E[\cdot]$ is the expectation value. An ARI of 1 denotes perfect alignment with the ground truth, while 0 indicates performance no better than random segmentation.

Furthermore, we quantify the rate of merge and split errors for predicted chambers using the Variation of Information (VI) metric \cite{meilua2007comparing}, which is an entropy-based measure of clustering quality. It is defined as
\begin{equation}
    \text{VI}(\mathbf{P},\mathbf{G}) = H(\mathbf{P}|\mathbf{G}) + H(\mathbf{G}|\mathbf{P}),
\end{equation}
where the term $H(\mathbf{P}|\mathbf{G})$ measures over-segmentation (split error, denoted $\mathrm{VI}_{\text{split}}$), reflecting how much the predicted segmentation breaks apart true objects from the ground truth; on the other hand, $H(\mathbf{G}|\mathbf{P})$ captures under-segmentation (merge error, denoted $\mathrm{VI}_{\text{merge}}$), which indicates how multiple ground truth regions are merged into fewer predicted objects. Lower values for these entropies indicate better segmentation performance.

\subsubsection{Chamber Ordering}
We use the Spearman's rank correlation coefficient ($\rho$) to assess the precision of chamber ordering. Only predicted chamber centroids that lie within ground truth chambers are included in ranking. After matching the centroids, we have two sets of ranks based on the labels of the centroids in the prediction and ground truth. These ranks are then used to calculate the Spearman's rank correlation coefficient, i.e.,
\begin{equation}
    \rho = 1 - \frac{6\sum_{k=1}^M r_k^2}{M(M^2-1)},
\end{equation}
where $r_k$ is the rank difference between matched chambers. The value of $\rho$ ranges from $-1$ to $1$. A value of $\rho=1$ indicates perfect agreement between the predicted and ground truth orders. A value of $\rho=-1$ indicates perfect disagreement. A value of $\rho=0$ suggests no correlation between the predicted and ground truth orders.

\subsubsection{Chamber Centroids}
We also employ the Euclidean distance ($\delta$) to measure the spatial proximity between the predicted and ground truth centroids of individual chambers. Let $(x_j^{(p)}, y_j^{(p)}, z_j^{(p)})$ and $(x_j^{(g)}, y_j^{(g)}, z_j^{(g)})$ denote the predicted and ground truth coordinates of the centroid of chamber $S_j$, respectively. The Euclidean distance between these centroids is computed for each matched pair, and the average across all $M$ chambers in the dataset is used as an evaluation metric, i.e.,
\begin{equation}
\delta = \frac{1}{M} \sum_{j=1}^{M} \|(x_j^{(p)}, y_j^{(p)}, z_j^{(p)}) - (x_j^{(g)}, y_j^{(g)}, z_j^{(g)})\|_F,
\end{equation}
Smaller values of $\delta$ indicate closer alignment of predicted and true chamber centroids. In contrast, larger values of $\delta$ indicate greater discrepancies between the predicted and ground truth centroids.

\subsection{Pipelines Comparison}
All quantitative and qualitative results reported in this section are based on the test set comprising 15 samples. The mean values of each metric across samples are summarized in Table~\ref{tab:metrics}. Results on the training set are omitted, as they exhibit similar trends and do not provide additional insights. To ensure comparability, 2D method outputs were rescaled to match the spatial resolution of 3D methods before evaluation.

\begin{figure*}[!htbp]
    \centering
    \subfloat[]
    {\includegraphics[width=0.45\linewidth]{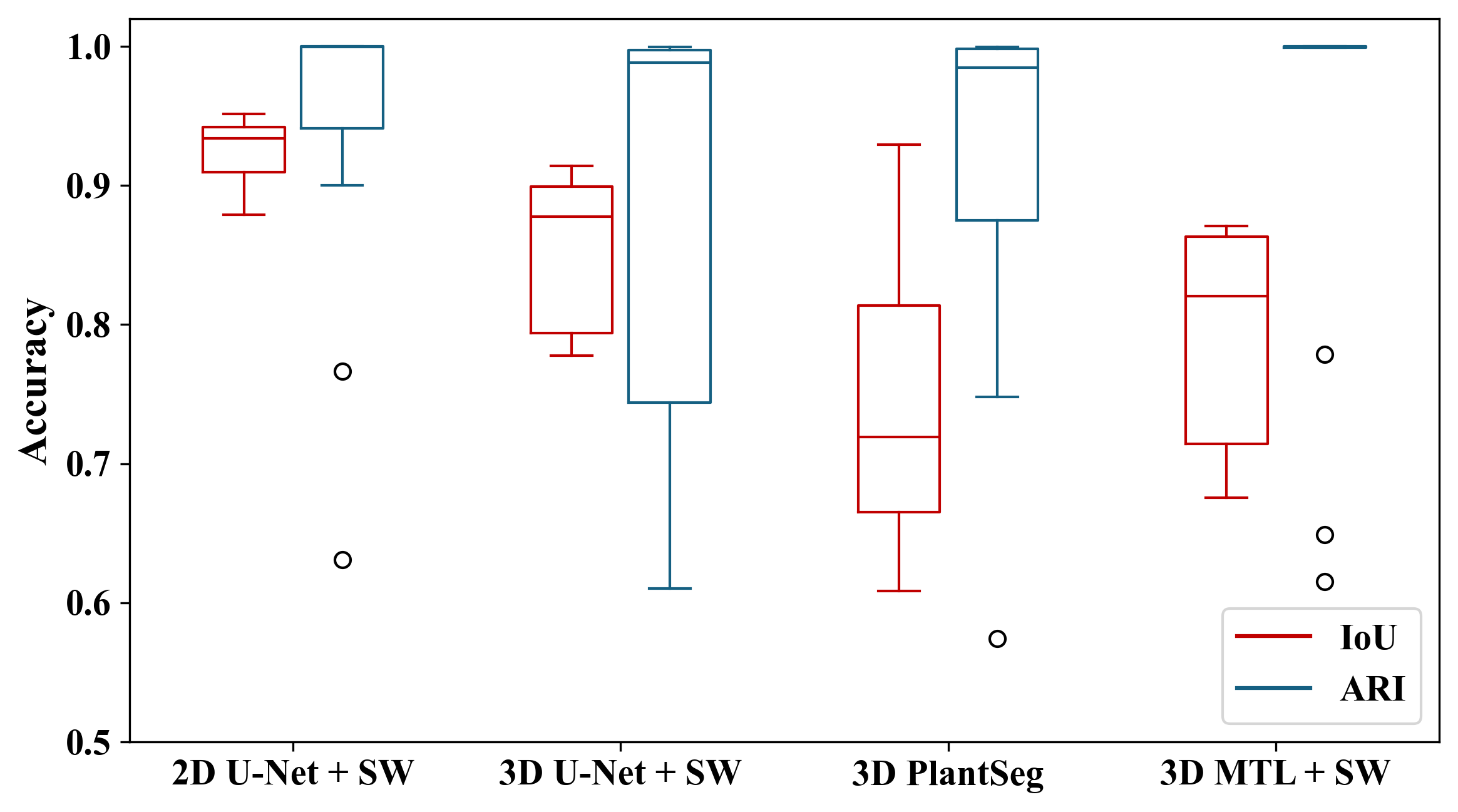}%
    \label{fig:iou_ari}}
    \hfil
    \subfloat[]
    {\includegraphics[width=0.45\linewidth]{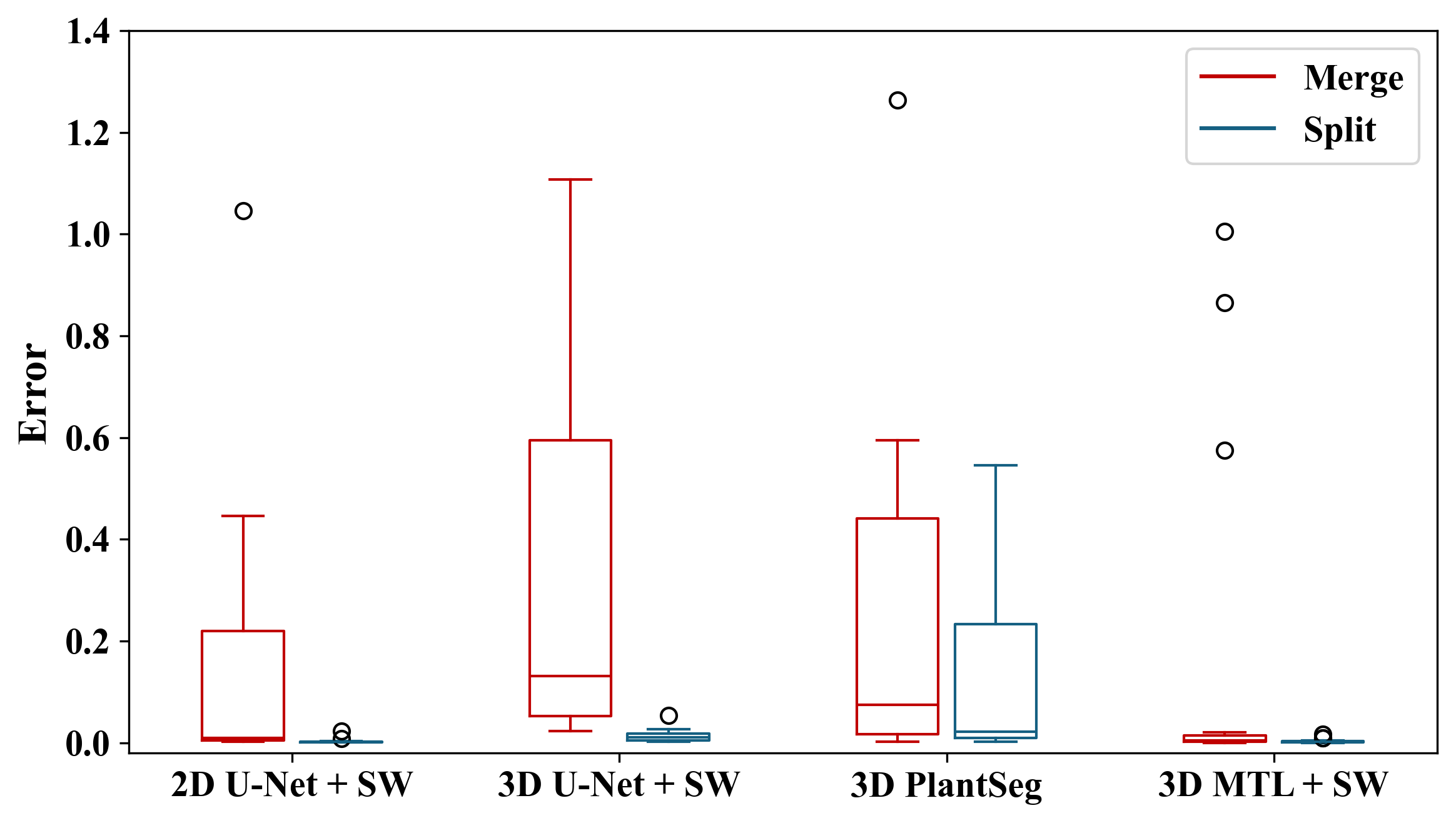}%
    \label{fig:merge_split}}\\
    \subfloat[]{\includegraphics[width=\linewidth]{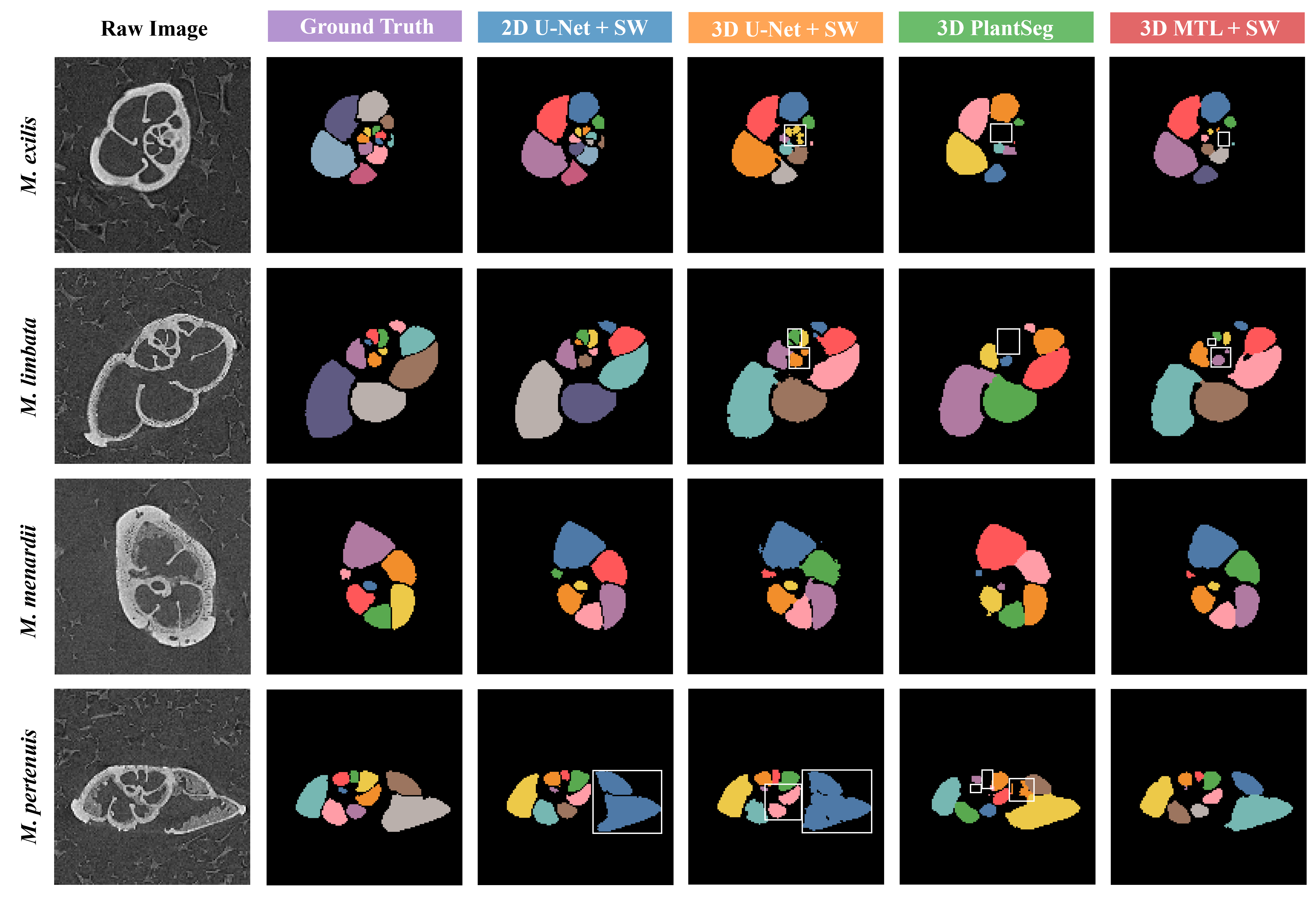}%
    \label{fig:instance_seg}}
    \caption{Evaluation of predicted instance segmentation compared to ground truth on the test set, along with representative slices of instance segmentation results for four species of \textit{Menardella}. (a) shows the segmentation accuracy of different chamber segmentation methods on the test set. While the 2D U-Net + SW achieved the highest performance in overall chamber segmentation based on IoU computed on binary masks, the 3D MTL + SW showed comparable performance in instance-level segmentation when evaluated using the ARI on instance masks. (b) shows the segmentation errors, where all segmentation methods tend to under-segment chambers. (c) presents example instance segmentation results by different methods for each species. The first column shows raw CT image slices, the second column shows the ground truth chamber segmentation, and the remaining columns display predictions from the tested methods. Different chamber instances were randomly assigned different colors, while white bounding boxes highlight regions where the methods made errors due to over- or under-segmentation or missing chambers.}
    \label{fig:instance_seg}
\end{figure*}

\subsubsection{Segmentation Performance}
\begin{table}[]
\centering
\caption{Mean values of evaluation metrics for each segmentation method evaluated on the test set. Metrics include Intersection over Union (IoU), Adjusted Rand Index (ARI), Variation of Information for merge and split errors (VI$_\mathrm{merge}$, VI$_\mathrm{split}$), number of valid detected chambers ($M$), Spearman’s correlation coefficient ($\rho$), and Euclidean distance ($\delta$).}
\scalebox{0.95}{
\begin{tabular}{cccccccc}
\toprule
\textbf{Pipelines}     & ${\mathrm{IoU}}$  & ${\mathrm{ARI}}$  & ${\mathrm{VI}_{\text{merge}}}$ & ${\mathrm{VI}_{\text{split}}}$ & ${M}$ & ${\rho}$ & ${\delta}$ \\ \midrule
2D U-Net + SW & 0.93 & 0.95 & 0.16        & 0.00        & 14               & 0.96          & 1.51      \\
3D U-Net + SW & 0.85 & 0.89 & 0.35        & 0.01        & 9                & 0.97          & 1.64      \\
3D PlantSeg   & 0.75 & 0.92 & 0.26        & 0.09        & 9                & 0.87          & 1.72      \\
3D MTL + SW   & 0.79 & 0.94 & 0.17        & 0.00        & 10               & 0.96          & 0.49      \\ \bottomrule
\end{tabular}
}
\label{tab:metrics}
\end{table}

We evaluated the segmentation accuracy of each method using IoU and ARI as shown in Fig.~\ref{fig:instance_seg}a, which capture overall segmentation performance and the consistency of instance-level clustering, respectively. Among all the segmentation methods, the 2D U-Net + SW achieved the highest performance in terms of the IoU metric, reaching a median value of 0.93, followed by 3D U-Net + SW and then the 3D MTL + SW, both yielding median IoU values within the range of 0.80 to 0.90. The 3D PlantSeg, in particular, exhibited the lowest performance, with a median IoU of 0.72. Although the 3D segmentation methods show a notable discrepancy in IoU compared to the 2D U-Net + SW, their performance is comparable when evaluated using ARI ($\approx$ 0.9-1.0), indicating broadly consistent instance-level labeling.

To further interpret the over- and under-segmentation reflected in the ARI scores, we analyzed the merge and split errors, as shown in Fig.~\ref{fig:instance_seg}b. All segmentation methods exhibited higher merge errors than split errors, with the greatest discrepancy observed in the 3D U-Net + SW ($\mathrm{VI}_{\text{merge}}=0.35$ vs $\mathrm{VI}_{\text{split}}=0.01$). This indicates that our methods tend to preserve chamber connectivity while being conservative about creating false boundaries. The 3D MTL + SW reduced merge errors the most ($\mathrm{VI}_{\text{merge}}=0.17$) compared to other 3D methods, validating that explicit learning both interior and boundary features improves boundary discrimination. Sample instance segmentation results for each method are presented in Fig.~\ref{fig:instance_seg}c, where the diversity of segmentation behaviors is illustrated.

\subsubsection{Evaluation of Chamber Ordering}
\begin{figure}[!htbp]
    \centering
        \includegraphics[width=\linewidth]{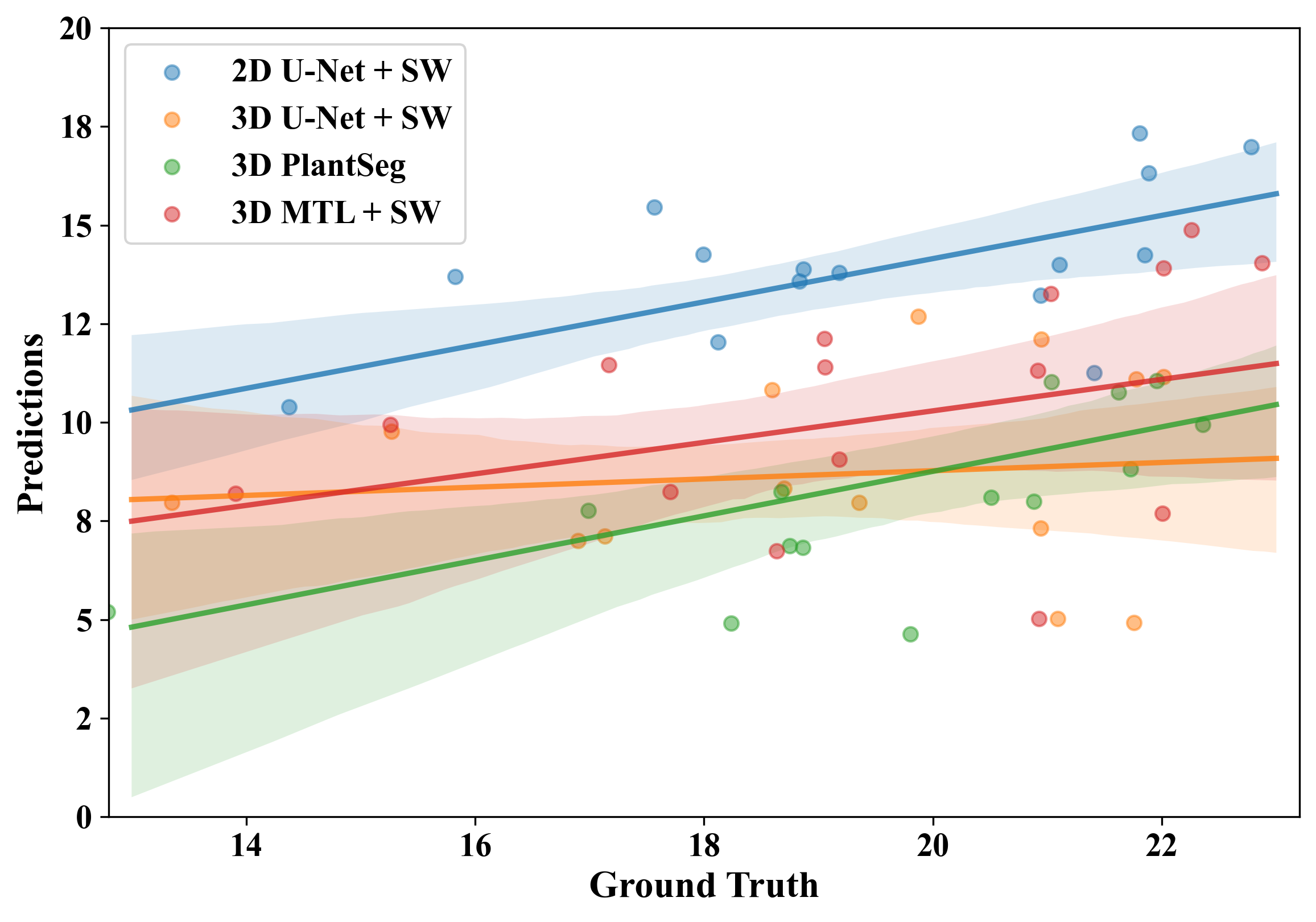}
    \caption{Regression analysis of predictions versus ground truth for measuring the number of chambers on the test set. The scatter plot with jitter shows the relationship between predicted values and actual measurements. Each line represents the fitted linear regression, and the shaded area indicates the 95\% confidence interval around the regression line.}
    \label{fig:num_chambers}
\end{figure}

We evaluated the accuracy of chamber ordering considering chamber counts, sequence correctness, and centroid localization. Based on the ordinary least squares (OLS) regression analysis in Fig.~\ref{fig:num_chambers}, the 2D U-Net + SW identified the largest number of chambers ($M=14$, $\approx$ 71\% of the ground truth) and exhibited the narrowest 95\% confidence interval around its regression line. The 3D U-Net + SW and 3D PlantSeg consistently identified approximately 9 chambers per sample, while the 3D MTL + SW achieved the highest mean chamber count among the 3D methods ($M=10$), capturing more than half of the ground truth chambers.

\begin{figure}[!htbp]
    \centering
        \includegraphics[width=\linewidth]{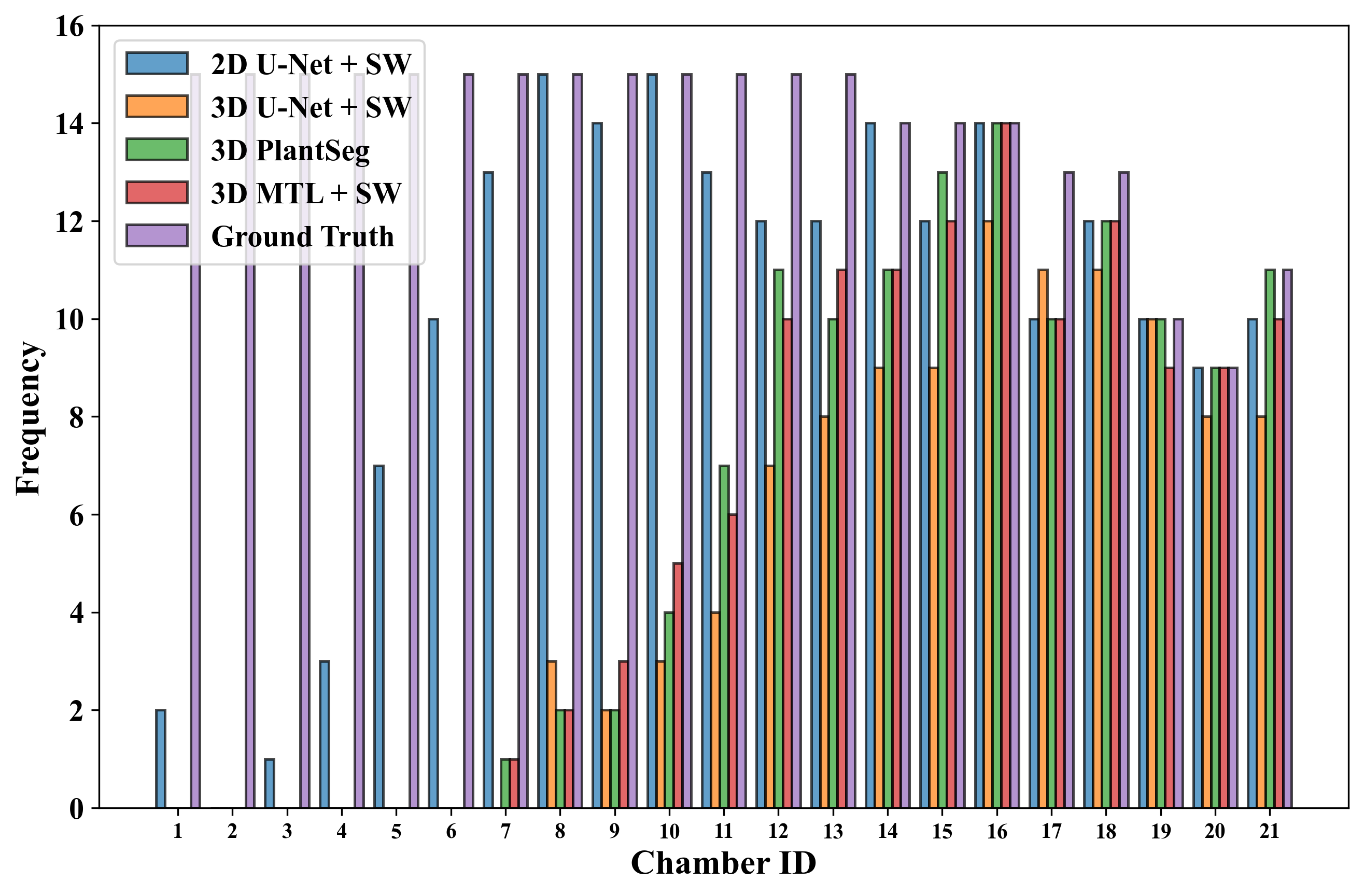}
    \caption{Distribution of chamber IDs detected by different segmentation methods and the ground truth on the test set. All segmentation methods exhibit a tendency to under-detect or miss smaller chambers, which correspond to lower chamber IDs.}
    \label{fig:chambers_id}
\end{figure}

To investigate the occurrence of missing chambers in more detail, we analyzed the distribution of detected chamber IDs in Fig. \ref{fig:chambers_id}. Chamber IDs follow the biological growth order in each sample, providing insight into which developmental stages each method captures most effectively. The 2D U-Net + SW was the only method that consistently captured both early-stage and well-developed chambers, while all 3D methods struggled with the first six chambers across samples. Although the 2D U-Net + SW captured some of these early chambers, it still fell short of matching the ground truth distribution.

\begin{figure*}[!htbp]
    \centering
    \subfloat[]
    {\includegraphics[width=0.45\linewidth]{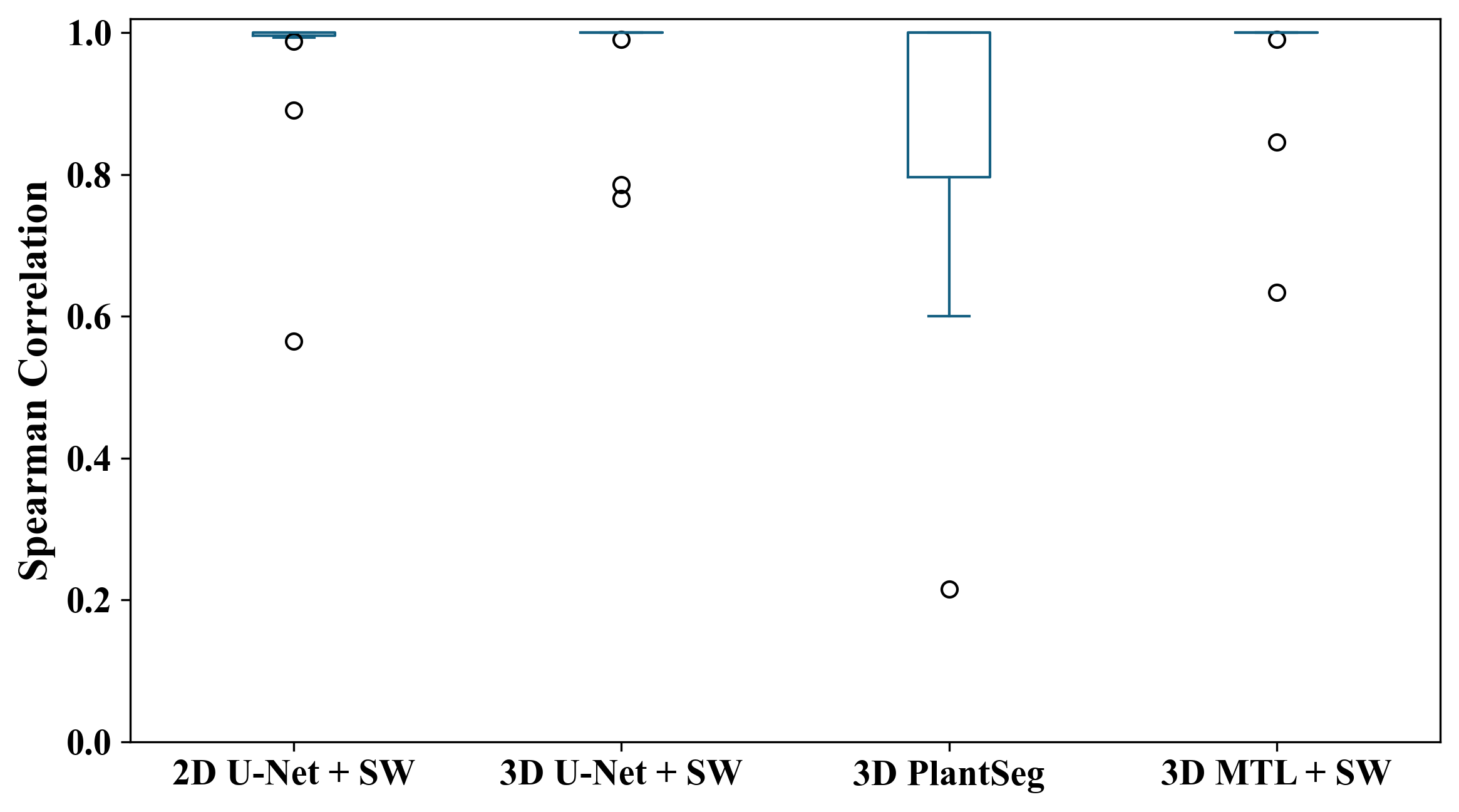}%
    \label{fig:spearman}}
    \hfil
    \subfloat[]
    {\includegraphics[width=0.45\linewidth]{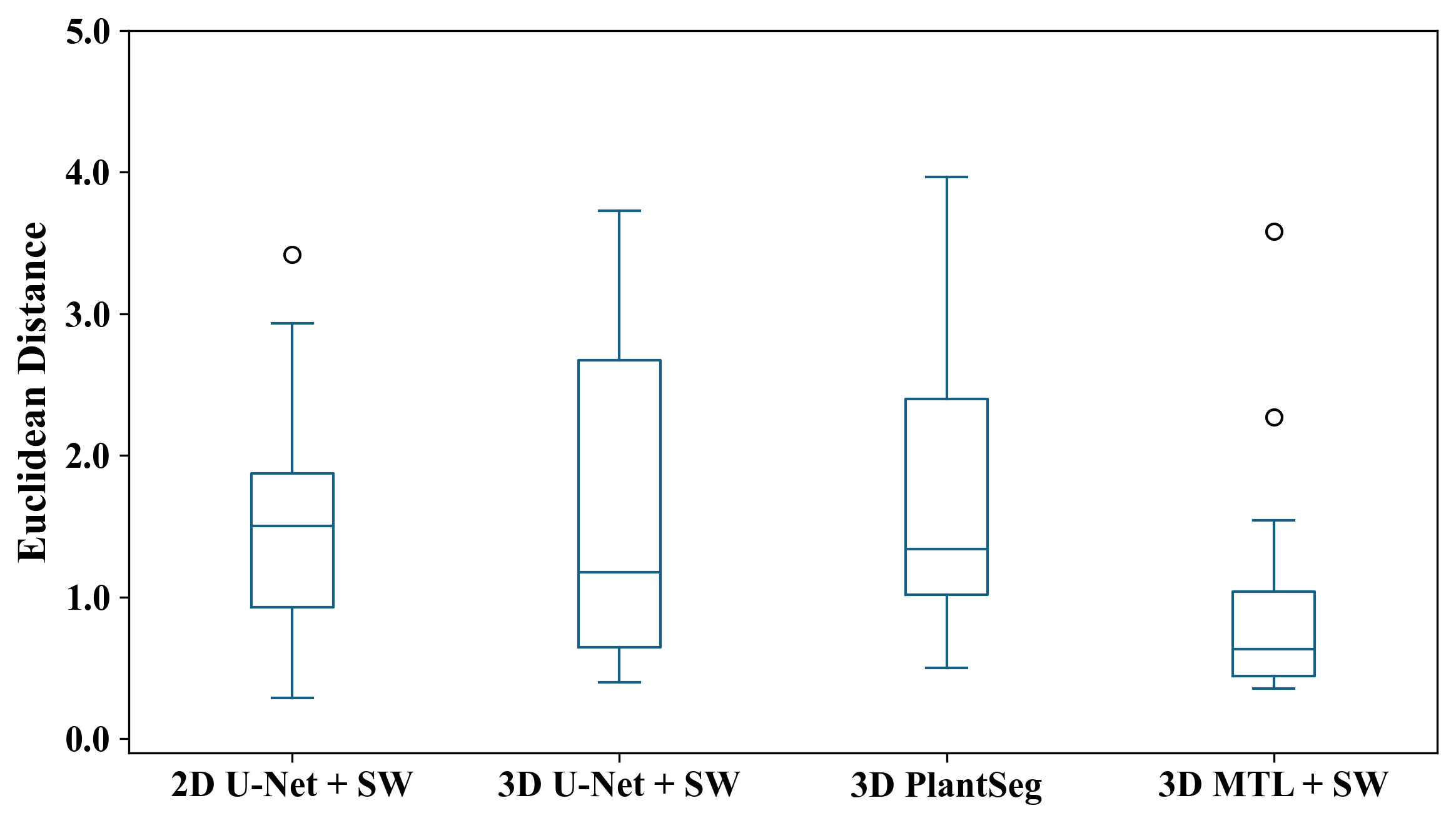}%
    \label{fig:distance}}\\
    \subfloat[]{\includegraphics[width=\linewidth]{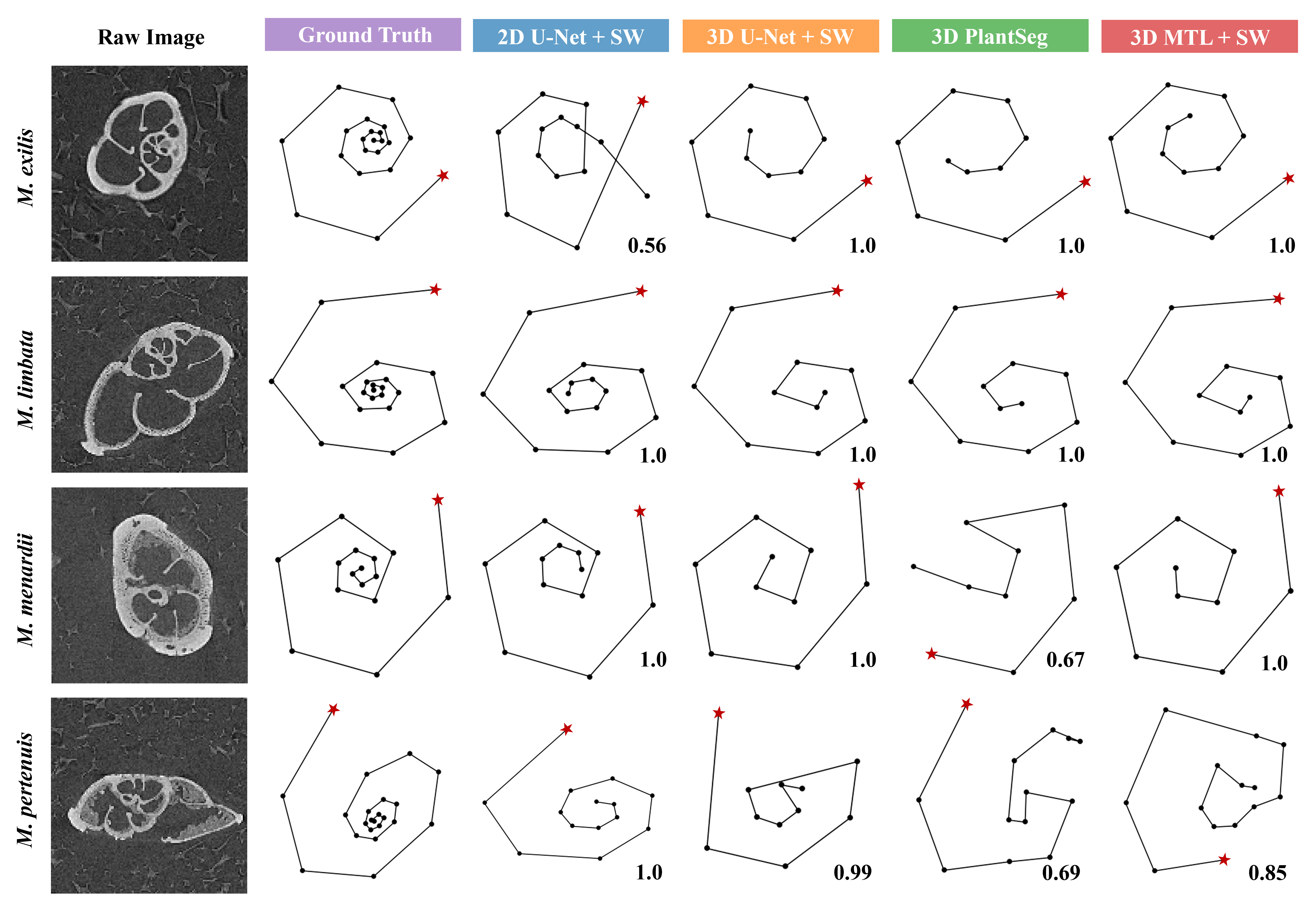}%
    \label{fig:growth_trajectory}}
    \caption{Evaluation of predicted chamber organization compared to ground truth on the test set, along with representative chamber ordering results for four species of \textit{Menardella}. (a) shows the Spearman’s rank correlation coefficient between the predicted chamber order and the ground truth order. All segmentation methods exhibit positive correlation values, indicating consistent alignment with the true chamber sequence, with the majority of predictions achieve a perfect correlation of 1. (b) shows the Euclidean distance between the predicted and ground truth centroids of each chamber. All segmentation methods demonstrate consistently low errors, with medium centroid distances around 1 voxel. (c) presents example growth trajectory results from different segmentation methods for each species, with the final chamber marked by a red star. The first column displays raw CT image slices, the second column shows the ground truth growth trajectories, and the remaining columns illustrate predictions from the tested segmentation methods. The Spearman’s correlation for each prediction is shown in the lower right corner of each panel.}
    \label{fig:chamber_ordering}
\end{figure*}

All proposed methods exhibited strong spatial understanding and ordering capability. As shown in Fig.~\ref{fig:chamber_ordering}a, the nearest-neighbor-based algorithm successfully ordered the detected chambers along their growth trajectory, with all methods achieving a high median Spearman correlation ($\rho \geq 0.87$). In Fig.~\ref{fig:chamber_ordering}b, all segmentation methods demonstrated high accuracy in predicting the spatial locations of chamber centroids, with errors below 4 voxels across all samples. Remarkably, the 3D MTL + SW achieved the highest centroid localization accuracy, with a median Euclidean distance of less than 1 voxel between the predicted and ground truth centroids across all chambers. Examples of chamber ordering results are presented in Fig.~\ref{fig:chamber_ordering}c.

\section{Discussion}
\label{sec:discussion}
In this study, we proposed an end-to-end pipeline for automated chamber ordering in planktonic foraminifera, built upon segmentation techniques from the field of computer vision. We explored multiple segmentation methods that emphasize different aspects of image features, including region-based, boundary-aware, and combined strategies. Specifically, the image data were first processed using U-Net-based architectures for semantic segmentation of binary ROIs, followed by hand-crafted post-processing algorithms to delineate individual chambers. Finally, the segmented chambers were ordered sequentially using a nearest-neighbor-based algorithm that leverages both chamber size and centroid information. Our experimental results on the existing \textit{Menardella} dataset demonstrate that our pipeline can ultimately achieve strong performance in the chamber ordering task. Notably, the ordering accuracy is consistently high regardless of the segmentation method employed, as evidenced by near-perfect median Spearman's correlation coefficients between predicted and ground truth chamber sequences. Furthermore, the strong spatial agreement between the predicted and ground truth chamber centroids (mean error: 1.47 voxels) confirms accurate recovery of three-dimensional growth trajectories.

Among the evaluated approaches, the 2D U-Net demonstrated superior performance in semantic segmentation (median IoU: 0.93) compared to 3D U-Net counterpart. This advantage stems from the input of high-resolution 2D slices, which preserve fine-grained structural details that are degraded in 3D methods due to volumetric downsampling and interpolation. Furthermore, treating each 2D slice as an independent training sample increases the effective data diversity and enhances generalization. Similar observations have been reported in other tasks \cite{baumgartner2018exploration, crespi20223d}. Despite the differences in semantic segmentation accuracy observed between 2D and 3D methods, the 2D U-Net + SW and the 3D MTL + SW exhibit comparable performance in terms of instance-level segmentation (mean ARI: $\approx$ 0.95), indicating that both approaches effectively mitigate over- and under-segmentation errors. The superior performance of the 2D method can be attributed to its higher input resolution, which expands the receptive field and enables more precise boundary localization. In parallel, the 3D MTL model benefits from multi-feature fusion in its encoder, enhancing its ability to distinguish chamber interior from boundaries despite voxel fidelity loss associated with 3D downsampling.

However, while the proposed pipeline can capture the gross morphology and order of chambers, it struggles to detect small, early-stage chambers. The evaluation results across all segmentation methods revealed a consistent tendency to merge adjacent chambers, as evidenced by the higher merge errors relative to split errors. This merging behavior leads to an underestimation of total chamber counts compared to the ground truth. One likely reason for this trend is the limited inter-chamber contrast and narrow sutural gaps, which pose challenges for learning-based methods to delineate precise boundaries. A similar tendency toward merging can also be observed when segmenting compact plant cells, as reported in \cite{kar2022benchmarking}. This ambiguity is further exacerbated in 3D volumetric data used in all 3D methods due to anisotropic resolution, where lower z-axis resolution hinders the model's ability to resolve fine structural separations. Our experiments further demonstrate that explicitly learning multiple feature representations yields more precise delineation. This effect is particularly evident in the 3D MTL + SW approach, which achieves lower merge errors compared to its 3D counterparts trained solely on chamber interior features (3D U-Net + SW) or boundary features (PlantSeg). In addition to merge errors, all segmentation methods also demonstrated a noticeable drop in detection frequency for smaller chambers, with this issue being more pronounced in the 3D methods. These chambers typically correspond to the earliest ontogenetic stages, which are smaller, denser, and more likely to be fused or occluded in CT data. Consequently, segmentation models either omit or incorrectly merge these structures due to reduced voxel fidelity and limited spatial separation.

Future work may benefit from incorporating boundary-aware training objectives, multi-scale attention mechanisms, or adaptive instance-level refinements to mitigate merge errors and enhance small object sensitivity. While the nearest-neighbor-based algorithm demonstrates strong performance in ordering chambers within the \textit{Menardella} genus, other genera may exhibit distinct chamber morphologies, growth patterns, and spatial arrangements \cite{lin20243dkmi}. Future research should therefore explore more flexible ordering frameworks that integrate geometric priors and biological constraints to generalize across diverse morphologies.

\section{Conclusion}
\label{sec:conclusion}
In conclusion, we proposed an end-to-end pipeline for the automated reconstruction of 3D chamber growth trajectories in chamber-secreting organisms such as planktonic foraminifera, enabling both quantitative and qualitative assessment of their developmental dynamics. The pipeline involves instance segmentation of individual chambers using a combination of deep learning and hand-crafted post-processing methods, followed by a nearest-neighbor-based ordering algorithm that arranges the chambers sequentially according to their spatial configuration and size, thereby reconstructing biologically plausible growth trajectories in 3D space. We systematically compared multiple segmentation strategies, including both 2D and 3D architectures emphasizing different feature domains. Although all segmentation variants achieved reliable chamber ordering, our findings underscore the importance of voxel resolution and multi-feature learning in volumetric instance segmentation, which jointly mitigate over- and under-segmentation in morphologically complex specimens. Overall, the proposed pipeline provides a scalable and objective framework for analyzing chamber formation and growth patterns in planktonic foraminifera, with strong potential to advance quantitative research in morphological evolution.

\section*{Acknowledgments}
The authors acknowledge the use of the Iridis high-performance computing facility at the University of Southampton in the completion of this work. This study was funded by the Natural Environment Research Council (NE/P019269/1). We acknowledge $\mu$-VIS X-ray Imaging Centre (https://muvis.org), part of the National Facility for laboratory-based X-ray CT (nxct.ac.uk—EPSRC: EP/T02593X/1) at the University of Southampton for the provision of the $\mu$CT imaging, processing, and data management infrastructure.

\bibliographystyle{IEEEtran}
\bibliography{main}

\end{document}